\documentclass[11pt]{article}

\usepackage[preprint]{acl}

\usepackage{times}
\usepackage{latexsym}

\usepackage[T1]{fontenc}
\usepackage[utf8]{inputenc}

\usepackage{microtype}

\usepackage{inconsolata}

\usepackage{graphicx}
\usepackage{amsmath}
\usepackage{amssymb}
\usepackage{bbold}
\usepackage[table]{xcolor}
\usepackage{array}
\usepackage{arydshln}
\usepackage{caption}
\usepackage{subcaption}
\usepackage{booktabs}
\usepackage{tabularx}
\usepackage{algorithm}
\usepackage{algpseudocode}
\usepackage{amsthm}

\definecolor{LightPurple}{HTML}{D6D6FF}

\usepackage[T1]{fontenc}
\usepackage[utf8]{inputenc}

\usepackage{microtype}

\usepackage{inconsolata}
\usepackage{amsmath}
\usepackage{amssymb}
\usepackage{booktabs}
\usepackage{comment}
\usepackage{subcaption}

\usepackage{graphicx}

\newcommand{\E}{\mathbb{E}}

\newcommand{\mc}[1]{\mathcal{#1}}

\newcommand{\topm}{\operatorname{TopM}}
\newcommand{\proj}{\operatorname{Proj}}

\newcommand{\ProtoAda}{\textsc{ProtoAda}}

\author{Yu-Cheng Shi\textsuperscript{* 2},
        Zhen-Hao Xie\textsuperscript{* 1,2},
        Jun-Tao Tang\textsuperscript{2},
        Da-Wei Zhou\textsuperscript{1,2,\textdagger} \\
        \textsuperscript{1} School of Artificial Intelligence, Nanjing University, China \\
        \textsuperscript{2} State Key Laboratory of Novel Software Technology, Nanjing University, China \\
        \textsuperscript{*} Equal contribution 
        \textsuperscript{\textdagger} Corresponding Author \\
        {\tt \small
        \{231250034,juntao.tang\}@smail.nju.edu.cn, 
        \{wenzh,zhoudw\}@lamda.nju.edu.cn
        }
}

\title{\ProtoAda: Prototype-Guided Adaptive Adapter Expansion and Geometric Consolidation
for Multimodal Continual Instruction Tuning}

\begin{document}
\maketitle
\begin{abstract}
Multimodal Large Language Models (MLLMs) achieve strong performance through instruction tuning, but real-world deployment requires them to continually acquire new vision-language capabilities, making Multimodal Continual Instruction Tuning (MCIT) essential. To reduce inter-task interference and promote collaboration, recent methods often employ sparse architectures like Mixture of LoRA Experts with image-text similarity routing. However, tasks with distinct response structures could share highly similar visual-linguistic semantics and thus be wrongly routed to the same expert; image-text similarity alone is insufficient for reliable task assignment. For example, an expert in a grounding task requiring coordinate prediction may be biased toward producing short textual answers after learning semantically similar VQA tasks. This format-blind task assignment integrates heterogeneous response types into shared parameters, inducing gradient interference and ineffective expert collaboration. To address this problem, we propose \ProtoAda, a prototype-guided adaptive tuning framework. \ProtoAda\ introduces format-aware task prototypes to align task assignment and routing with both task semantics and output structure, and further consolidates format-compatible updates in a geometry-aware manner to effectively reuse and progressively refine existing parameters. Extensive experiments on multiple benchmarks demonstrate that \ProtoAda\ achieves superior performance, especially on tasks whose answer structures are easily corrupted by sequential tuning.
\end{abstract}

\section{Introduction}
\begin{figure}[t]
    \centering
    \includegraphics[width=0.9\linewidth]{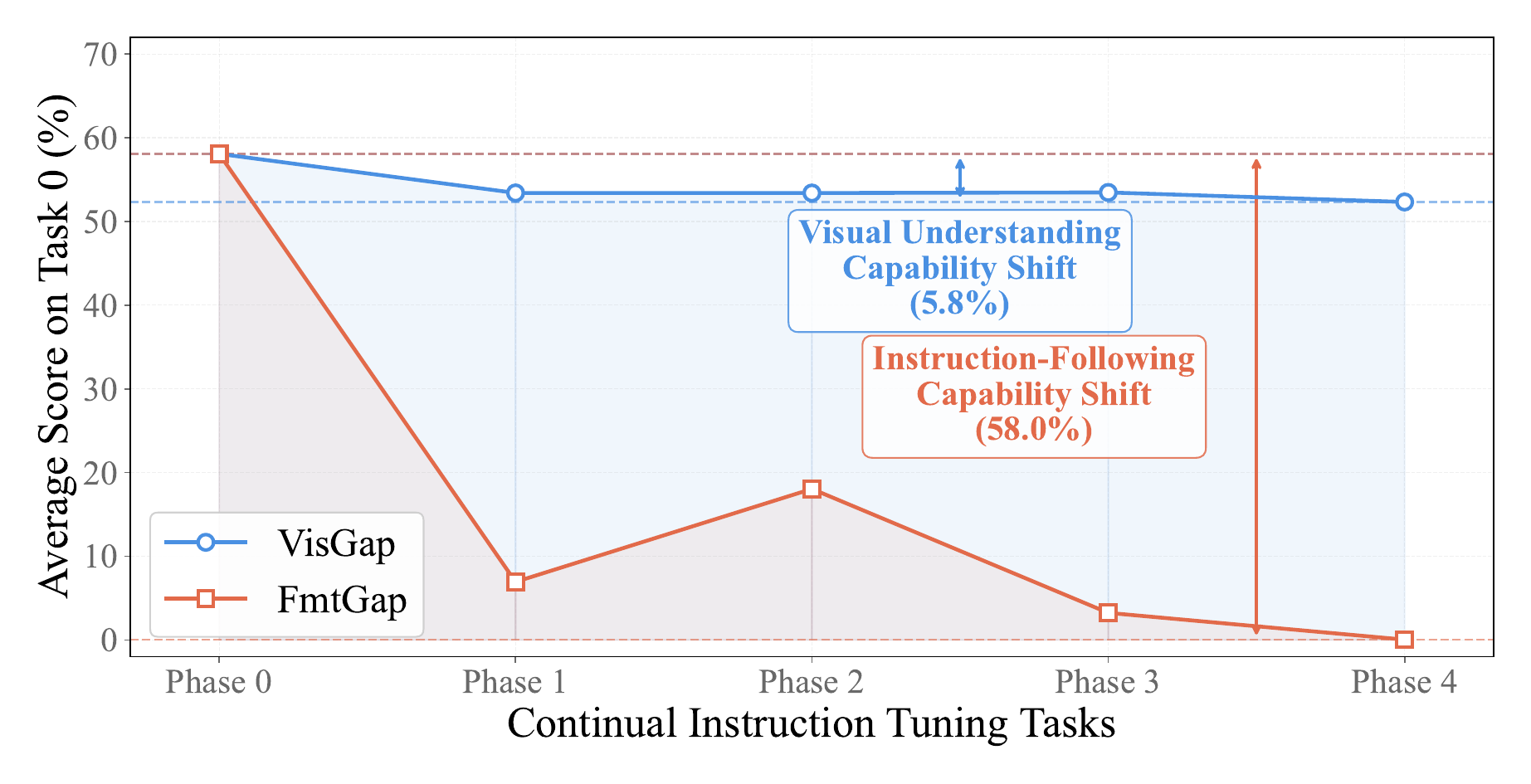}
    \caption{Accuracy of sequential finetuning under format-varied (FmtGap) and semantic-varied (VisGap) continual tuning streams.}
    \label{fig:pre_fmt_cor}
    \vspace{-4mm}
\end{figure}

\begin{figure}[t]
    \centering
    \begin{subfigure}[t]{0.49\linewidth}
		\includegraphics[width=1\linewidth]{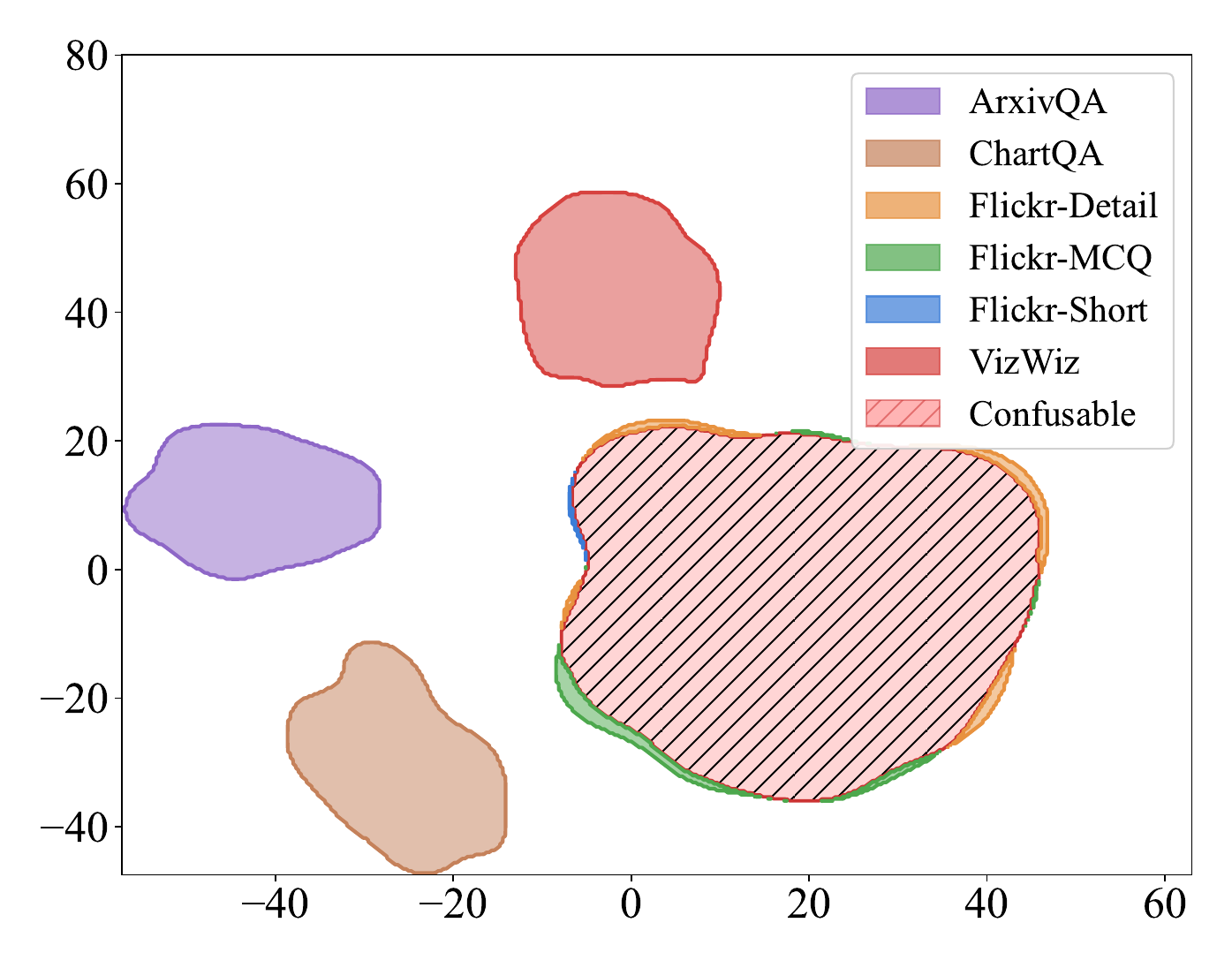}
		\caption{Semantic-Only.}
		\label{fig:latency}
	\end{subfigure}
	\hfill
	\begin{subfigure}[t]{0.49\linewidth}
		\includegraphics[width=1\linewidth]{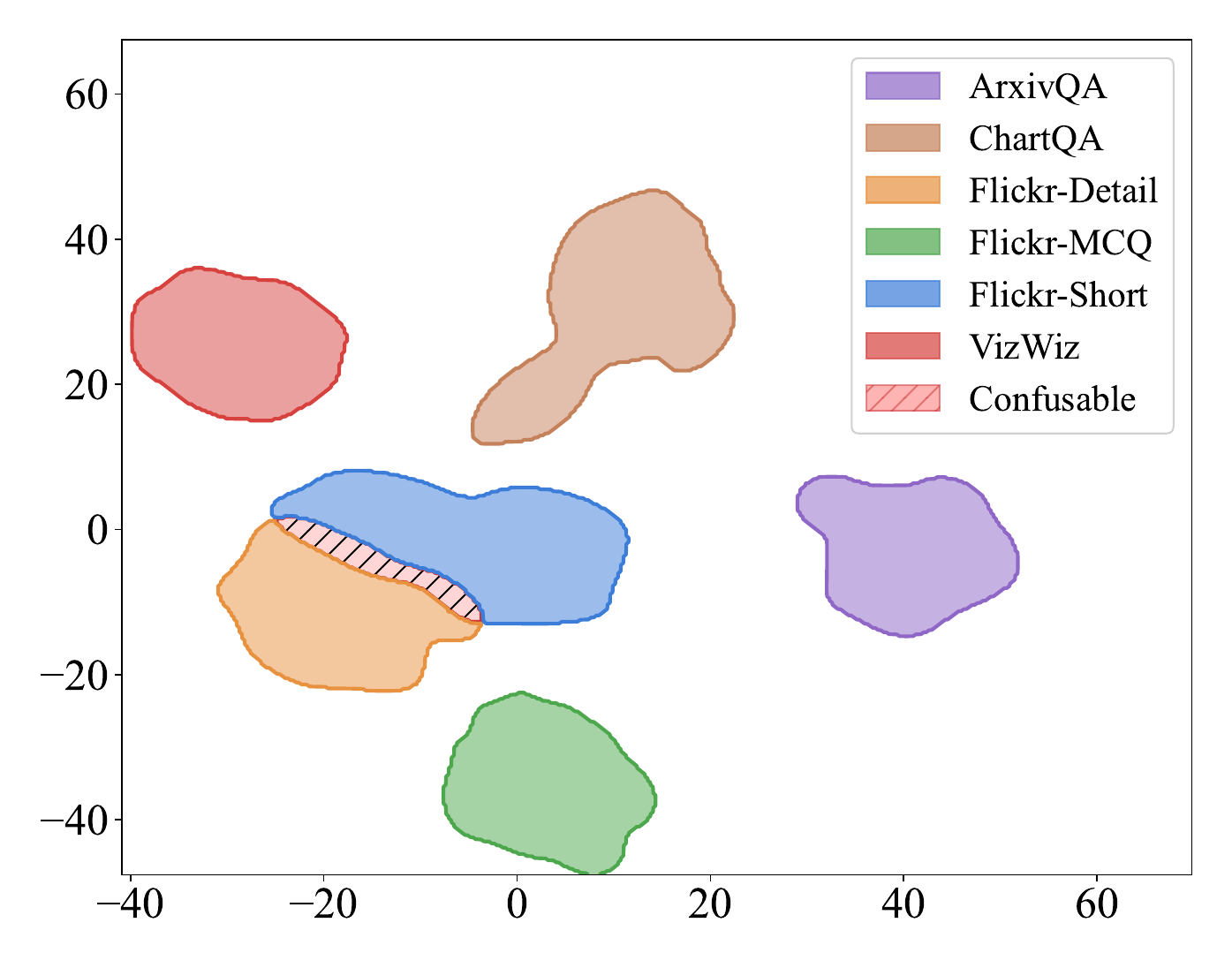}
		\caption{Format-Aware.}
		\label{fig:tradeoff}
	\end{subfigure}
    \caption{t-SNE visualization of task representations under semantic-only and format-aware prototypes.}
    \label{fig:pre_vis_fmt}
\end{figure}
Recently, multimodal Large Language Models (MLLMs)~\cite{Qwen-VL,liu2023llava,zhu2023minigpt} have shown strong generalization through multimodal instruction tuning~\cite{zhang2023instruction,tong2025metamorph} on large-scale datasets, enabling a wide range of vision-language tasks~\cite{radford2021learning,dai2023instructblip,guo2025mammoth}. In realistic scenarios, however, multimodal tasks~\cite{hu2021unit,yang2025magic} often arrive sequentially, requiring multimodal continual instruction tuning (MCIT)~\cite{chen2024coin} to continually acquire new capabilities after deployment, without access to previous training data. This setting is challenging due to catastrophic forgetting~\cite{french1999catastrophic,french2020modeling}, where learning new tasks may overwrite not only previously learned semantic knowledge but also task-specific output conventions~\cite{liu2025continual}. Recent works introduce parameter-efficient adaptation to improve the stability-plasticity trade-off~\cite{zeng2025modalprompt,yu2025progressive,zhao2025llava,huai2025cl,wang2025smolora}, with MoE-style parameter isolation mitigating interference by routing tasks to sparsely activated lightweight experts, typically based on image-text semantic similarity. However, semantic similarity alone cannot determine whether tasks should share the same adaptation path: tasks involving similar visual regions and instructions may require incompatible output protocols, such as bounding-box coordinates versus short textual phrases. Such mismatches can entangle heterogeneous formats within experts, motivating an explicit format-aware mechanism for MCIT.

Format-induced incompatibility suggests that the failure of semantic routing may arise from a deeper mismatch: semantically similar instructions can require fundamentally different response protocols. To verify this, we construct two controlled continual-learning streams under the same parameter-sharing setup. The first keeps the visual-semantic content fixed by using Flickr30k~\citep{plummer2015flickr30k}, but varies the response protocol across five formats~\footnote{The five formats are brief description, detailed description, short/one-word answer, multiple choice answer, and yes/no answer.}; the second keeps the output format as brief description, but mixes different datasets like Flickr30k and VizWiz~\citep{gurari2018vizwiz} to introduce visual-semantic shifts. As shown in Fig.~\ref{fig:pre_fmt_cor}, sequential tuning degrades performance in both streams, but the decline is substantially larger under format variation. This result indicates that MLLM tuning not only learns visual-linguistic associations but also aligns instructions with expected answer forms. Therefore, visual-linguistic similarity is insufficient for parameter-sharing decisions, and response-protocol compatibility should be explicitly considered.

This raises a practical question: can response formats be captured in a lightweight way before deciding task sharing? For each task, we construct two representations from frozen image-text embeddings: a semantic-only prototype and a format-aware prototype augmented with simple response statistics, such as average length and token entropy. The t-SNE~\citep{van2008visualizing} visualization in Fig.~\ref{fig:pre_vis_fmt} shows that semantic-only embeddings poorly separate tasks with different answer protocols, while the lightweight format code makes these protocols more distinguishable. This suggests that response statistics can serve as an effective proxy for instruction-following patterns, guiding task grouping to encourage sharing among compatible protocols while reducing interference between incompatible ones.

These observations suggest that effective MCIT should both identify incompatible response protocols to avoid format-specific conflicts and consolidate compatible updates for parameter reuse. To this end, we propose \ProtoAda, a prototype-guided adaptive tuning framework that builds format-aware task prototypes from visual, language, and response-format statistics, grouping tasks by both content and expected answer protocol. Within each group, \ProtoAda\ performs geometry-aware consolidation: group-aligned directions refine shared parameters, while residual directions are preserved to protect task-specific response conventions. Experiments on multiple benchmarks demonstrate the effectiveness of \ProtoAda\ on complex continual data streams.

\begin{figure*}
    \centering
    \includegraphics[width=\linewidth]{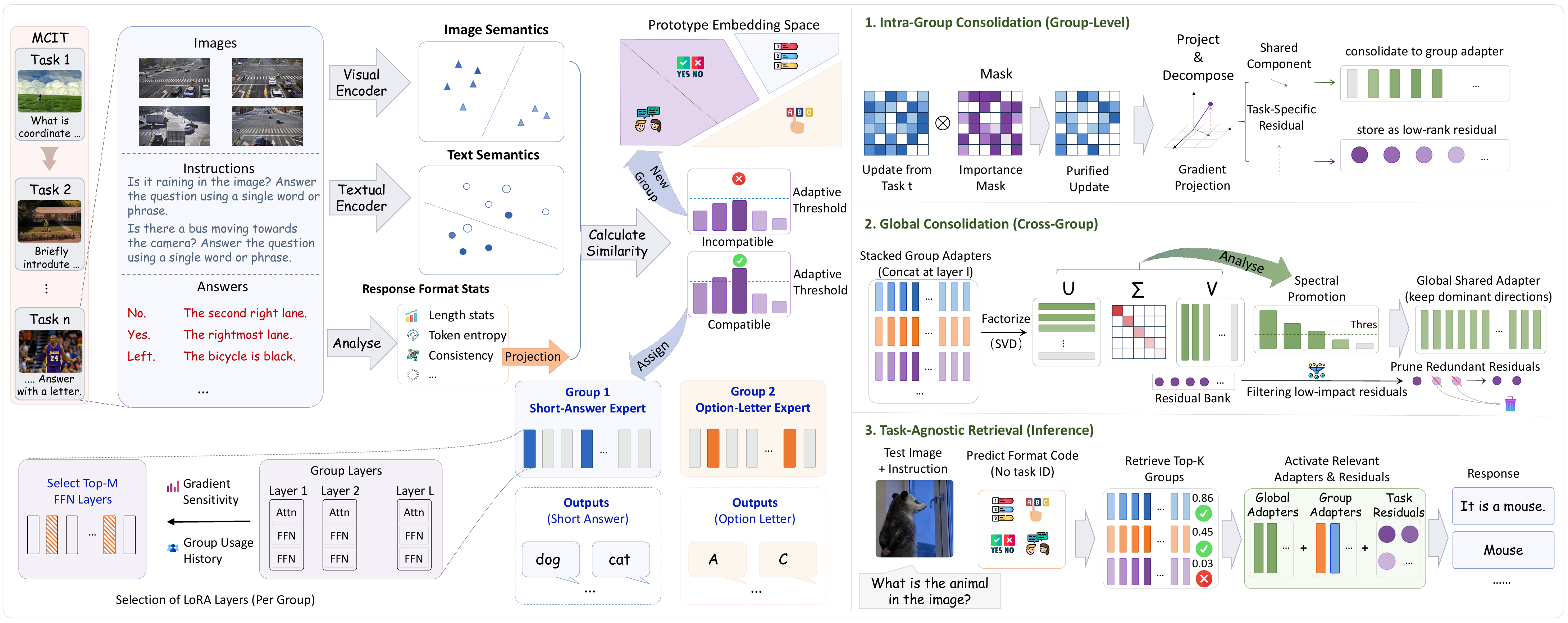}
    \caption{\ProtoAda\ groups tasks with compatible response formats into shared experts, while separating conflicting formats and adapting only the most sensitive layers within each group.}
    \label{fig:teaser}
\end{figure*}

\section{Related Work}
As MLLMs are increasingly deployed in open-world scenarios, continual instruction tuning~\citep{chen2024coin,liu2025continual} becomes essential for acquiring new capabilities without overwriting old ones. Existing methods mainly follow three directions. Replay-based methods~\citep{li2025multimodal,yu2025progressive} retain or synthesize previous image-text samples to stabilize later training, but the required storage and computation increase as the task stream grows~\citep{lee2025oasis}. Regularization- and prompt-based methods~\citep{wang2023orthogonal,cao2024continual,liu2025llava} constrain representation drift or introduce lightweight task-specific prompts, improving efficiency but often relying on semantic task similarity or task identity cues that can be unreliable when visually similar tasks require different answer formats~\citep{zeng2025modalprompt,chen2025sefe}. Parameter-efficient and expert-based methods~\citep{guo2025hide,zhao2025llava,huai2025cl,wang2025smolora} further introduce LoRA modules, adapter pools, or mixture-of-experts structures to isolate task-specific knowledge and reduce interference.

\section{Preliminaries}

\subsection{Multimodal continual instruction tuning}

We consider a multimodal large language model consisting of a frozen vision encoder $\phi$, a multimodal projector $\pi$, and a large language model $f$ equipped with lightweight trainable modules. Let $\{\mc{T}_1,\mc{T}_2,\ldots,\mc{T}_N\}$ denote a sequence of multimodal instruction-tuning tasks, where each $\mc{T}_i=\{(v_{j},q_{j},y_{j})\}_{j=1}^{M_i}$ contains an image $v_{j}$, an instruction $q_{j}$, and a target response $y_{j}=(y_{j,1},\ldots,y_{j,L_j})$. The frozen vision encoder extracts visual features $\mathbf{\tilde v}=\phi(v)$, the embedding layer $\psi(\cdot)$ produces instruction embeddings $\mathbf{u}=\psi(q)$, and the projector maps visual features into the language space as $\mathbf{w}=\pi(\mathbf{\tilde v})$, forming the multimodal input $\mathbf{z}=[\mathbf{w};\mathbf{u}]$. Given a target response token sequence $y=(y_1,\dots,y_L)$, the MLLM models the conditional distribution:
\begin{equation}
\label{eq:mcit_ar}
    p_{\Theta}(y\mid z)=\prod_{k=1}^{L_j}
    p_{\Theta}(y_{k}\mid \mathbf{z},y_{<k}).
\end{equation}
The goal of MCIT is to obtain a unified model that performs well on all tasks observed so far. After step $i$, the ideal objective is
\begin{equation}
\label{eq:mcit_objective}
    \Theta_i^* =
    \arg\min_{\Theta}
    \E_{(v,q,y)\sim \mc{T}_{\le i}}
    \left[
    -\log p_\Theta(y\mid \mathbf{z})
    \right],
\end{equation}
where $\mc{T}_{\le i}$ denotes the union of all seen task distributions. In rehearsal-free MCIT, however, optimization at step $i$ only accesses $\mc{T}_i$, and thus the learner must improve the current task likelihood while preserving behaviors induced by previous tasks without revisiting their samples. This mismatch of data distribution is the fundamental source of forgetting.

\subsection{Low-rank adaptation in the language model}

We adopt a rehearsal-free parameter-efficient setting in which the vision encoder, projector, and pretrained language backbone are frozen, and trainable parameters are introduced only through LoRA modules inserted into all linear modules of the language model. For a linear module $\ell$ with frozen weight $W_\ell$, LoRA modifies its output as
\begin{equation}
\label{eq:lora_module}
    \mathbf{o}_\ell = \mathbf{W}_\ell \mathbf{h}_\ell + \mathbf{B}_\ell \mathbf{A}_\ell \mathbf{h}_\ell,
\end{equation}
where $\mathbf{A}_\ell\in\mathbb{R}^{r\times d_{\mathrm{in}}}$ and $\mathbf{B}_\ell \in\mathbb{R}^{d_{\mathrm{out}}\times r}$ are trainable low-rank factors with rank $r$. Only $\Theta=\{\mathbf{A}_\ell ,\mathbf{B}_\ell \}_{\ell\in\mc{L}}$ are updated. After learning task $t$, we denote the LoRA increment as $\Delta\Theta_t=\Theta_t-\Theta_{t-1}$, which must adapt to the current task while avoiding disruption to previously learned instruction-following behaviors. LoRA-based continual instruction tuning thus requires careful decisions about which parameter group receives a new task and how its update is consolidated.
\paragraph{Discussions.}
The gap between the ideal objective in Eq.~\eqref{eq:mcit_objective} and rehearsal-free training makes LoRA updates shown in Eq.~\eqref{eq:lora_module} vulnerable to structural interference: when task allocation relies mainly on image-instruction semantics, tasks with similar inputs but incompatible answer formats may share the same LoRA group, corrupting the instruction-to-protocol mapping. This motivates \ProtoAda, which represents tasks with visual, language, and format-aware prototypes before sharing parameters, and consolidates compatible updates to enable reliable reuse.

\section{Method}
\label{sec:method}
To preserve instruction-following ability for MLLMs, we propose \ProtoAda, a prototype-guided MCIT framework. \ProtoAda\ organizes LoRA parameters into multiple memory groups, where each group stores shared parameters for tasks that are compatible in both image-text semantics and response protocols. To avoid entangling incompatible instruction formats, when a new task arrives, \ProtoAda\ builds a format-aware task prototype from visual semantics, language semantics, and response-format statistics. The task is then assigned to an existing group or a new group according to its compliance with old tasks inferred from its prototype. Compatible updates are then consolidated into the corresponding group, enabling parameter reuse while preserving task-specific output conventions.

\subsection{Format-aware task grouping}
\label{subsec:format_aware_grouping}
To better represent tasks with prototypes and more effectively identify compatibility relations among tasks, given a task $\mc{T}_i=\{(v_{t,j},q_{t,j},y_{t,j})\}_{j=1}^{M_i}$, we first compute its language and visual descriptors using frozen visual and textual encoders such as CLIP~\citep{radford2021learning}:
\begin{equation}
\label{eq:text_vis_proto}
\begin{aligned}
&\mathbf{p}_t^{\mathrm{text}} = \frac{1}{M_t}\sum_{j=1}^{M_t} E_{\mathrm{text}}(q_{t,j}),\\
&\mathbf{p}_t^{\mathrm{vis}} = \frac{1}{M_t}\sum_{j=1}^{M_t} E_{\mathrm{vis}}(v_{t,j}).
\end{aligned}
\end{equation}
However, these descriptors only capture what the task is about, but not how the model should answer. As shown in Fig.~\ref{fig:pre_fmt_cor}, tasks with similar image-text semantics may require different response protocols. Routing such tasks only by image-text similarity can therefore force incompatible instruction formats into the same parameters and damage the model's instruction-following behavior. To incorporate response protocols into task grouping, we calculate the mean, variance of response length, token uncertainty, and template consistency~\footnote{Implementation details about the calculation of these format statistics are shown in Appendix.~\ref{app:fmt}} from target responses and derive a format code:
\begin{equation}\label{eq:format_code}
\mathbf{c}_t^{\mathrm{fmt}} = [\mu_{\ell}, \sigma_{\ell}, H_{\mathrm{tok}}, C_{\mathrm{fmt}}],
\end{equation}
This format code is mapped into the prototype space by:
\begin{equation}\label{eq:format_proto}
\mathbf{p}_t^{\mathrm{fmt}} = \mathbf{W}_{\mathrm{fmt}}\mathbf{c}_t^{\mathrm{fmt}}+\mathbf{b}_{\mathrm{fmt}}.
\end{equation}
The final task prototype combines language, vision, and format information:
\begin{equation}\label{eq:task_proto}
\mathbf{q}_t = \mathrm{Norm} \left( \mathbf{W}_q [\mathbf{p}_t^{\mathrm{text}}\|\mathbf{p}_t^{\mathrm{vis}}\|\mathbf{p}_t^{\mathrm{fmt}}] \right).
\end{equation}
Based on the format-aware task prototype, each memory group $g$ is represented by a group prototype $p_g$ that summarizes its assigned tasks. When a new task arrives, we compare its prototype $q_t$ with all existing group prototypes to decide whether it can share an existing group or should initialize a new one:
\begin{equation}\label{eq:group_similarity}
s_{t,g} = \frac{\langle \mathbf{q}_t,\mathbf{p}_g\rangle}
{\|\mathbf{q}_t\|_2\|\mathbf{p}_g\|_2},
\qquad
g_t=\arg\max_g s_{t,g}.
\end{equation}
Since different groups may exhibit different degrees of internal diversity, we adopt a dynamic admission threshold. Specifically, for each group $g$, we maintain the running mean $\mu_g$ and standard deviation $\sigma_g$ of the similarities between previously assigned tasks and the group prototype $\mathbf{p}_g$. The assignment rule is then defined as:
\begin{equation}\label{eq:adaptive_assignment}
\mathrm{Assign}(t)=
\begin{cases}
g_t, & s_{t,g_t}\ge \mu_{g_t}-\sigma_{g_t},\\
\mathrm{new}, & s_{t,g_t}< \mu_{g_t}-\sigma_{g_t}.
\end{cases}
\end{equation}
If the task is assigned to an existing group, its prototype is updated as:
\begin{equation}\label{eq:group_proto_update}
\mathbf{p}_{g_t} \leftarrow \mathrm{Norm}
\big((1-\alpha)\mathbf{p}_{g_t}+\alpha \mathbf{q}_t\big).
\end{equation}
Otherwise, a new group is initialized with $\mathbf{q}_t$. This allows compatible tasks to share LoRA parameters while separating tasks with conflicting response protocols.
However, since the format is unavailable during inference, we therefore train a lightweight predictor $f_\mathrm{fmt}$ to infer it from input-side descriptors during training:
\begin{equation}\label{eq:format_predictor}
 \mathbf{\hat c}_t^{\mathrm{fmt}} = f_{\mathrm{fmt}}
(\mathbf{p}_t^{\mathrm{text}}).
\end{equation}
During inference, we replace $\mathbf{c}^{\mathrm{fmt}}$ with the predicted format code $\mathbf{\hat c}^{\mathrm{fmt}}$ when constructing the task prototype, thereby enabling format-aware grouping under task-agnostic retrieval.

\noindent\textbf{Discussions.} By incorporating response protocols into task identity, Eq.~\eqref{eq:task_proto} fuses visual, linguistic, and format cues, while Eq.~\eqref{eq:adaptive_assignment} admits a task into a group only when it is compatible with the group's historical prototype. The format predictor further extends this principle to task-agnostic inference, where target responses are unavailable.

\subsection{Prototype-conditioned adaptive growth}

Once a task has been assigned to a memory group, \ProtoAda\ further decides what additional parameters should be introduced and where they should be placed. Adding Lora Modules to all layers for every task is parameter-inefficient and may perturb stable behaviors, so we selectively adapt only the layers where the current task demands the most change. For each FFN layer $\ell$, we estimate task sensitivity by the gradient energy of the task loss with respect to the hidden state:
\begin{equation}
\label{eq:layer_sensitivity}
    \gamma_\ell
    =
    \E_{(v,q,y)\sim\mc{T}_i}
    \left[
    \left\|
    \frac{\partial \mc{L}_{\mathrm{ce}}}{\partial \hat{h}_\ell}
    \right\|_2
    \right].
\end{equation}
A larger $\gamma_\ell$ indicates that layer $\ell$ is more involved in fitting the current task and is therefore a better candidate for adaptation.

To avoid repeatedly allocating capacity to only a few high-gradient layers, we further incorporate the historical layer-usage pattern of the assigned memory group. Since tasks in the same group share compatible instruction-following behaviors, adapting a historically used layer is more likely to reuse an established adaptation path rather than open an unrelated one. We thus combine current-task sensitivity with group-level usage:
\begin{equation}
\label{eq:layer_selection}
\begin{aligned}
    &\tilde{\gamma}_\ell
    =
    \gamma_\ell+\eta u_{g_t,\ell},\\
    &\mc{S}_t
    =
    \topm(\{\tilde{\gamma}_\ell\}_{\ell=1}^{L};M),
\end{aligned}
\end{equation}
where $u_{g_t,\ell}$ records how frequently group $g_t$ has used layer $\ell$, and $\mc{S}_t$ is the set of Linear layers selected for adaptation. Only the modules in $\mc{S}_t$ are updated for the current task.

\noindent\textbf{Discussions.}
This mechanism refines the memory assignment made by the prototype: the prototype decides which group hosts the task, while the sensitivity-and-usage score decides where new capacity is introduced within that group. The usage term provides a soft reuse prior for compatible tasks, while the sensitivity term still allows new layers to be selected when the current task demands them, balancing parameter reuse and task-specific expansion under format-aware grouping.

\subsection{Geometry-aware consolidation and task-agnostic retrieval}
\label{subsec:consolidation_retrieval}

After task $t$ is assigned to group $g_t$ and optimized, the LoRA increment $\Delta\Theta_t$ contains two types of information: group-compatible knowledge that benefits future tasks in the same group, and task-specific corrections that only preserve the current task's response behavior. Merging the whole update into the group adapter would entangle these two components and let fragile response protocols interfere with shared abilities. We therefore decouple the reusable part of each update from its task-specific residual before consolidation.

Since not every changed parameter is equally supported by the current task, we first compute an activation-aware importance score:
\begin{equation}
\label{eq:activation_importance}
    I_t(i,j)=|\Delta\Theta_t(i,j)|\,a_j,
\end{equation}
where $a_j=\|\mathbf{X}_{:,j}\|_2$ is the input activation norm of channel $j$. Keeping the top-$k\%$ entries yields a purified update~\citep{sun2024simple}:
\begin{equation}
\label{eq:purified_update}
    \Delta \hat{\Theta}_t=\mathbf{M}_t\odot\Delta\Theta_t,
\end{equation}
which is then decomposed according to the geometry of the assigned group. 
Let $\mc{U}_{g_t}$ denote the subspace spanned by the current shared memory of group $g_t$, and let 
$\proj_{\mc{U}_{g_t}}(\cdot)$ be the orthogonal projection onto this subspace. 
The component aligned with the group memory is absorbed into the shared adapter, while the orthogonal component is kept as a compact task-specific residual:
\begin{equation}
\label{eq:share_res_decompose}
\begin{aligned}
    &\Delta\Theta_t^{\mathrm{share}}
    =
    \proj_{\mc{U}_{g_t}}(\Delta \hat{\Theta}_t),
    \\
    &\Delta\Theta_t^{\mathrm{res}}
    =
    \Delta\hat{\Theta}_t-\Delta\Theta_t^{\mathrm{share}} .
    \end{aligned}
\end{equation}
Here, $\Delta\Theta_t^{\mathrm{share}}$ captures directions reusable by the assigned group, whereas $\Delta\Theta_t^{\mathrm{res}}$ preserves task-specific deviations outside the group subspace.

The shared component updates the memory of the assigned group with a prototype-dependent weight. 
Let $\mathbf{A}_{g}$ denote the shared adapter parameters maintained by group $g$. 
We update the selected group as:
\begin{equation}
\label{eq:group_absorb}
\begin{aligned}
    &\mathbf{A}_{g_t}
    \leftarrow
    \mathbf{A}_{g_t}
    +
    \lambda_{t,g_t}
    \Delta\Theta_t^{\mathrm{share}},\\
    &\lambda_{t,g}
    =
    \lambda_0
    \frac{\exp(\beta s_{t,g})}
    {\sum_{g'}\exp(\beta s_{t,g'})}.
    \end{aligned}
\end{equation}
So tasks closer to the group prototype contribute more strongly. The residual is compressed by truncated singular value decomposition~\citep{klema1980singular} as:
\begin{equation}
\label{eq:residual_svd}
    \mathbf{R}_t=\operatorname{SVD}_{r_{\mathrm{res}}}(\Delta\Theta_t^{\mathrm{res}}),
\end{equation}
which preserves task-specific corrections without storing a dense task copy. To avoid redundant memory growth, we periodically reorganize the memory hierarchy. After every $K$ newly observed tasks, the group adapters at layer $\ell$ are concatenated as
\begin{equation}
\label{eq:all_groups}
    W_{\mathrm{all},\ell}
    =
    [\mathbf{A}_{\ell,1}\|\mathbf{A}_{\ell,2}\|\cdots\|\mathbf{A}_{\ell,G}],
\end{equation}
and their dominant spectral directions are promoted into a global shared adapter:
\begin{equation}
\label{eq:spectral_promotion}
    \mathbf{A}_\ell^{\mathrm{shared}}
    \leftarrow
    \mathbf{A}_\ell^{\mathrm{shared}}
    +
    \Delta \mathbf{A}_\ell^{\mathrm{global}}.
\end{equation}
Residuals already explained by shared or group memory are then removed from the residual bank according to their coverage ratio
\begin{equation}
\label{eq:residual_coverage}
    \rho_t
    =
    \frac{
    \left\|
    \proj_{\mc{U}_{\mathrm{shared}}\cup \mc{U}_{g_t}}(\mathbf{R}_t)
    \right\|_F
    }
    {\|\mathbf{R}_t\|_F}.
\end{equation}
During inference, since the task identity is unavailable, we construct a query prototype using the predicted format code and retrieve the top compatible groups with similarity-based weights
\begin{equation}
\label{eq:inference_weight}
    w_g
    =
    \frac{\exp(s_g/\tau)}
    {\sum_{g'\in\mathrm{TopK}}\exp(s_{g'}/\tau)},
\end{equation}
and activates only residuals whose stored prototypes are close to the query. The prediction is produced by augmenting the frozen backbone with the global adapter, the retrieved group-level adapters, and the selected task residuals. This enables task-specific knowledge to be organized and retrieved within a unified format-aware prototype space.

\subsection{Summary of \ProtoAda}
\label{subsec:summary}

Overall, \ProtoAda\ runs as a unified MCIT pipeline based on a set of memory groups. For each incoming task, we first build a format-aware prototype from frozen visual and textual descriptors together with a format code (Eqs.~\eqref{eq:text_vis_proto}--\eqref{eq:task_proto}), which is extracted from target responses during training and predicted by a lightweight predictor at inference (Eq.~\eqref{eq:format_predictor}). The prototype is then compared with existing group prototypes (Eqs.~\eqref{eq:group_similarity}--\eqref{eq:group_proto_update}), so that the task either joins a compatible group or seeds a new one. Within the assigned group, LoRA modules are inserted only at layers that are both sensitive to the current task and historically used by the group (Eqs.~\eqref{eq:layer_sensitivity}--\eqref{eq:layer_selection}), and are optimized by a cross-entropy loss together with an MSE loss for the format predictor, while the backbone and prior adapters stay frozen. The resulting update is then purified by activation-aware masking and split into a group-compatible component and an orthogonal residual (Eqs.~\eqref{eq:activation_importance}--\eqref{eq:share_res_decompose}); the former is absorbed into the group adapter, and the latter is compressed via truncated SVD(Eqs.~\eqref{eq:group_absorb}--\eqref{eq:residual_svd}). Periodic spectral promotion further extracts common directions into the global adapter and prunes redundant residuals (Eqs.~\eqref{eq:all_groups}--\eqref{eq:residual_coverage}). At inference, the predicted format code yields a query prototype that retrieves the most compatible groups and residuals (Eq.~\eqref{eq:inference_weight}), and the frozen backbone augmented with the global, group, and selected residual adapters produces the response in a single forward pass, all within one shared format-aware prototype space.

\section{Experiments}
\label{sec:experiments}

\subsection{Experimental Setup}
\label{subsec:exp_setup}

\noindent\textbf{Benchmarks.}
We evaluate \ProtoAda\ on the following two benchmarks:
TriGap~\citep{xie2026same} introduces challenges through longer task sequences, broader domain shifts, and imbalanced data scales, encompassing ten tasks: PMCVQA~\citep{zhang2023pmcvqa}, DocVQA~\citep{mathew2021docvqa}, ChartQA~\citep{masry2022chartqa}, IconQA~\citep{lu2021iconqa}, InfographicVQA~\citep{mathew2022infographicvqa}, ArxivQA~\citep{li2024multimodal}, Roadside~\citep{guan2026roadscenevqa}, ChemVQA~\citep{sabando2020chemva}, FloodNetVQA~\citep{rahnemoonfar2021floodnet}, and CLEVR-Math~\citep{lindstrom2022clevr}.
UCIT~\citep{guo2025hide} employs strict filtering to mitigate pre-training contamination and comprises six tasks: ArxivQA~\citep{li2024multimodal}, CLEVR-Math~\citep{lindstrom2022clevr}, IconQA~\citep{lu2021iconqa}, ImageNet-R~\citep{hendrycks2021many}, VizWiz-Caption~\citep{gurari2018vizwiz}, and Flickr30k~\citep{plummer2015flickr30k}. 

\noindent\textbf{Evaluation metrics.}
Following \citet{zhou2024class,chen2024coin}, we denote by $\mathcal{A}_{s,t}$, the performance on task $s$ evaluated after training up to task $t$, with $T$ total tasks. We summarize the average final performance by $\Bar{\mathcal{A}}=\frac{1}{T}\sum_{s=1}^{T}\mathcal{A}_{s,T}$.

\begin{table*}[t]
\centering
\renewcommand{\arraystretch}{1.0}
\setlength{\extrarowheight}{1pt}
\caption{Performance on TriGap. The best and second-best results are highlighted in \textbf{bold} and \underline{underline}, respectively.}
\label{tab:trigap}
\resizebox{\linewidth}{!}{
\begin{tabular}{l|cccccccccc|c}
\hline
\rowcolor{gray!20}
\textbf{Methods} & PMCVQA & DocVQA & ChartQA & IconQA & InfographicVQA & ArxivQA & Roadside & ChemVQA & FloodNetVQA & CLEVR & Average \\
\hline

Zero-shot~\citep{liu2023llava} & 35.40 & 12.68 & 9.36 & 19.27 & 5.06 & 53.77 & 7.40 & 5.30 & 47.41 & 20.37 & 21.60 \\
\rowcolor{gray!10}
FT-LoRA~\citep{hu2022lora} & 34.20 & 23.32 & 9.84 & 37.07 & 23.53 & 83.83 & 7.00 & 12.70 & 80.31 & 60.27 & 37.21 \\
Replay-LoRA & 33.70 & 33.95 & 14.00 & 46.67 & 28.97 & 75.57 & 9.40 & 15.90 & 73.81 & 58.80 & 39.08 \\
\rowcolor{gray!10}
MoE-LoRA~\citep{chen2024coin} & 39.03 & 37.49 & 12.44 & 43.43 & 35.17 & 90.90 & 7.93 & 20.70 & \textbf{90.41} & \textbf{67.00} & 44.45 \\
HiDe-LLaVA~\citep{guo2025hide} & 37.00 & 33.20 & 10.52 & 41.97 & 24.09 & 79.20 & 7.73 & 11.17 & 57.39 & 23.00 & 32.53 \\
\rowcolor{gray!10}
ModalPrompt~\citep{zeng2025modalprompt} & 38.23 & 38.23 & 11.92 & 44.73 & 37.37 & 84.47 & 10.13 & 12.43 & 71.52 & 52.50 & 40.15 \\
CL-MoE~\citep{huai2025cl} & 40.53 & 36.79 & 13.72 & 52.70 & 32.27 & \underline{93.00} & 7.77 & 18.33 & 80.09 & \underline{65.90} & 44.11 \\
\rowcolor{gray!10}
SAME~\citep{xie2026same} & 41.60 & \textbf{43.87} & 17.56 & \underline{64.03} & \textbf{39.57} & 90.46 & \underline{10.83} & 21.77 & \underline{81.09} & 54.50 & 46.53 \\
DISCO~\citep{guo2025federated} & \underline{42.03} & \underline{43.50} & \underline{18.01} & 63.13 & \underline{38.23} & 91.27 & \underline{11.02} & \underline{22.13} & 80.25 & 55.87 & \underline{46.54} \\\hdashline
\rowcolor{gray!10}
\textbf{\ProtoAda\ (Ours)} & \textbf{42.87} & 43.01 & \textbf{18.96} & \textbf{66.03} & 36.31 & \textbf{93.50} & \textbf{12.97} & \textbf{23.73} & 74.03 & 57.87 & \textbf{47.23} \\
\hline
\end{tabular}}
\end{table*}
\begin{table*}[t]
\centering
\setlength{\extrarowheight}{1pt}
\setlength{\tabcolsep}{17pt}
\renewcommand{\arraystretch}{1.0}
\caption{Average performance of different methods on the UCIT benchmark. The best and second-best results are highlighted in \textbf{bold} and \underline{underline}, respectively.}
\label{tab:ucit_main}
\resizebox{\linewidth}{!}{
\begin{tabular}{l|cccccc|c}
\hline
\rowcolor{gray!20}
\textbf{Methods} & ImageNet-R & ArxivQA & Vizcap & IconQA & CLEVER & Flickr30k & Average \\
\hline

Zero-shot~\citep{liu2023llava} & 18.88 & 52.62 & 38.75 & 21.25 & 21.12 & 41.44 & 32.34 \\
\rowcolor{gray!10}
FT-LoRA~\citep{hu2022lora} & 29.33 & 55.30 & 45.51 & 26.13 & 13.07 & \underline{58.07} & 37.90 \\
MoE-LoRA~\citep{chen2024coin} & 58.43 & 77.57 & 44.83 & 68.90 & 56.73 & \textbf{58.27} & 60.79 \\
\rowcolor{gray!10}
Replay-LoRA & 76.93 & 87.07 & 54.31 & 56.43 & 36.40 & 55.94 & 61.18 \\

CL-MoE~\citep{huai2025cl} & 64.12 & 78.38 & 44.83 & 62.00 & 50.75 & 58.06 & 59.69 \\
\rowcolor{gray!10}
HiDe-LLaVA~\citep{guo2025hide} & 87.62 & 91.12 & 42.68 & 57.62 & 31.00 & 50.41 & 60.08 \\
ModalPrompt~\citep{zeng2025modalprompt} & 80.50 & 90.62 & \underline{60.13} & 63.50 & 55.75 & 57.09 & 67.93 \\
\rowcolor{gray!10}
DISCO~\citep{guo2025federated} & 88.88 & \textbf{94.25} & 47.52 & 69.50 & 60.75 & 56.32 & 69.54 \\
SAME~\citep{xie2026same} & \textbf{89.91} & 91.40 & 55.33 & \underline{77.51} & \textbf{68.85} & 55.43 & \underline{73.07} \\\hdashline
\rowcolor{gray!10}
\textbf{\ProtoAda\ (Ours)} & \underline{89.83} & \underline{93.20} & \textbf{61.49} & \textbf{81.37} & \underline{66.40} & 55.64 & \textbf{74.66}\\
\hline
\end{tabular}
}
\end{table*}  
\noindent\textbf{Compared Methods.}
We compare \ProtoAda\ with state-of-the-art MCIT methods, including zero-shot evaluation~\citep{liu2023llava}, standard FineTune~\citep{hu2022lora}, MoELoRA~\citep{chen2024coin}, Replay-LoRA, HiDe-LLaVA~\citep{guo2025hide}, CL-MoE~\citep{huai2025cl}, ModalPrompt~\citep{zeng2025modalprompt}, DISCO~\citep{guo2025federated}, and SAME~\citep{xie2026same} . Zero-shot results serve as a lower-bound reference. To ensure a fair comparison, all methods use identical backbone architectures and data protocols.

\noindent\textbf{Implementation details.}
All experiments are conducted on 4 NVIDIA RTX 5090 GPUs. We follow Prism~\citep{tang2026prism} to conduct all experiments.
Following \citet{chen2024coin,xie2026same}, we use LLaVA-v1.5-7B~\citep{liu2023llava} as the backbone MLLM and CLIP-L/14-336~\citep{radford2021learning} to extract visual and textual features. The vision encoder, projector, and pretrained language model are frozen, and trainable low-rank modules are inserted into all linear layers of the language model~\citep{wang2025loki,zhu2025teach}. Each task is trained for 1 epoch using the AdamW optimizer with a learning rate of $2e-4$ and a linear warm-up ratio of 0.03. Unless otherwise specified, the initial rank is set to $r_0=8$ for each task in TriGap and $r_0 = 16$ for each task in UCIT. For \ProtoAda, the residual rank is $r_{\mathrm{res}}=2$, the number of selected FFN layers is $M=24$, and spectral memory maintenance is performed every $K=5$ tasks. The format predictor is a two-layer MLP trained jointly with the task loss.

\subsection{Benchmark Comparison and Ablation}
\label{subsec:exp_results}

\noindent\textbf{Benchmark Comparison.}
We compare \ProtoAda\ with representative continual instruction tuning baselines on TriGap and UCIT, as reported in Tab.~\ref{tab:trigap} and Tab.~\ref{tab:ucit_main}. On the more challenging TriGap benchmark, \ProtoAda\ achieves the best average accuracy of $47.23\%$, showing its robustness under longer task sequences and more heterogeneous response protocols. Consistent gains on UCIT further validate the effectiveness of our format-aware and consolidation-aware design, which enables compatible LoRA sharing while preserving task-specific instruction-following behaviors across benchmarks.

\noindent\textbf{Ablation Study.}
Tab.~\ref{tab:ablation_ucit} studies the contribution of each component in our method. We use MoELoRA as the baseline and fix the LoRA rank to 8 for all variants. Format-aware prototypes improve the average accuracy from $60.79\%$ to $68.13\%$ by grouping tasks with compatible response formats. Adaptive assignment further increases accuracy to $72.35\%$ through more flexible memory selection. Method with Geometry-aware consolidation (full \ProtoAda) achieves the best result of $74.66\%$ by preserving task-specific update residuals while sharing common directions.

\begin{table}[t]
\centering
\footnotesize
\setlength{\tabcolsep}{5pt}
\renewcommand{\arraystretch}{1.0}
\caption{Ablation study of \ProtoAda, all experiments are conducted under UCIT.}
\label{tab:ablation_ucit}
\resizebox{\linewidth}{!}{
\begin{tabular}{l|c}
\hline
\rowcolor{gray!20}
\textbf{Variant} & \textbf{Avg. Acc (\%)} \\
\hline
Baseline & 60.79 \\
\rowcolor{gray!10}
w/ Format & 68.13 \\
w/ Format + Adaptive & 72.35 \\
\rowcolor{gray!10}
w/ Format + Adaptive + Geometry (Ours) & \textbf{74.66} \\
\hline
\end{tabular}}
    \vspace{-4mm}
\end{table}

\subsection{Further Analysis}
\label{subsec:controlled_format_stream}
\begin{figure}[t]
\centering
\includegraphics[width=\linewidth]{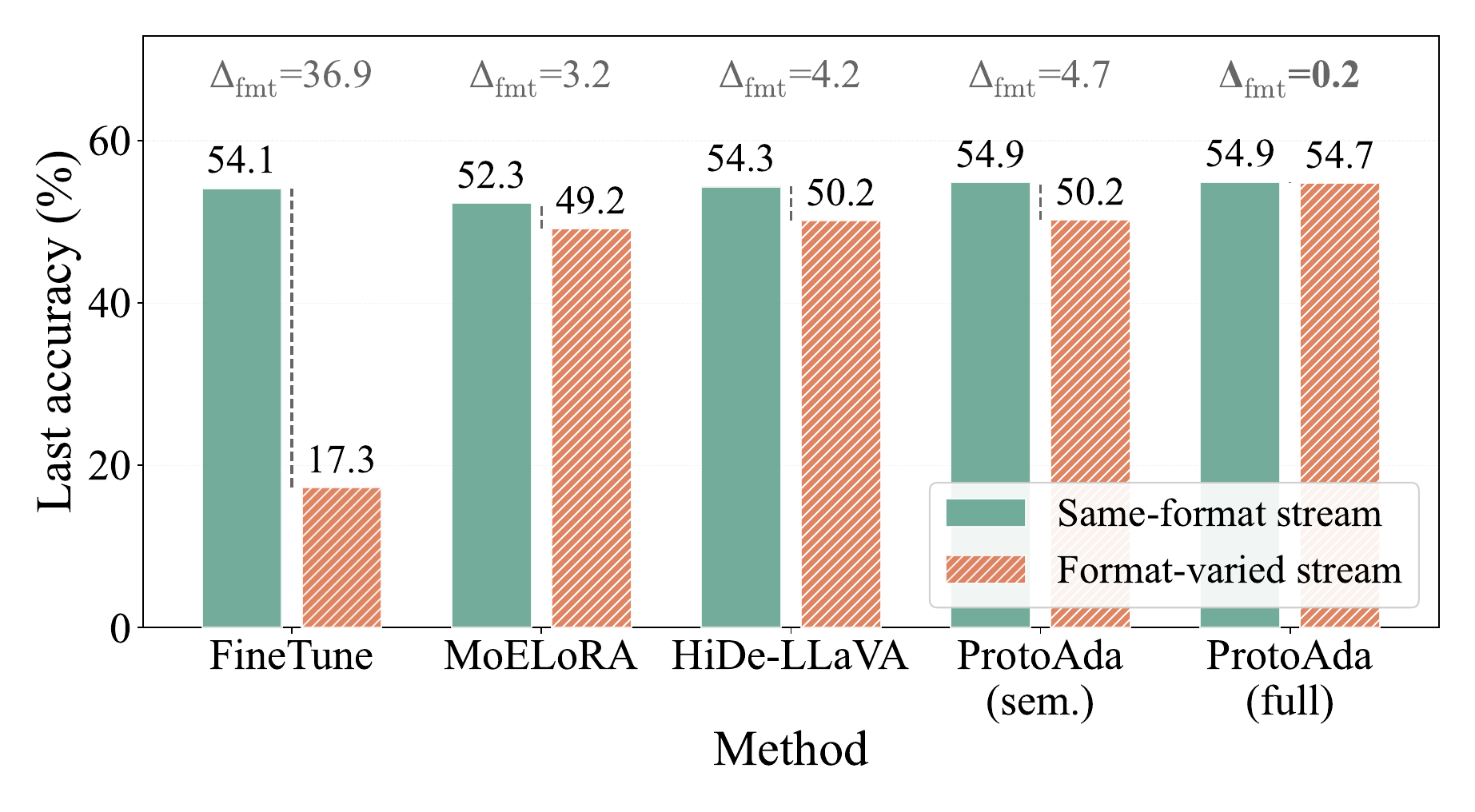}
\caption{Performance gap between handling streams with the same format and streams with varied formats.}
\label{fig:format_stress}
    \vspace{-5mm}
\end{figure}

\noindent\paragraph{Stress Test on Controlled Format Streams.}
To verify whether \ProtoAda\ addresses the failure mode identified in Fig.~\ref{fig:pre_fmt_cor}, we replay the two controlled streams from the preliminary study and compare full continual learners. As shown in Fig.~\ref{fig:format_stress}, all methods achieve similar Last accuracy on the same-format stream, mostly around $52$--$55\%$, indicating that the response protocol itself does not cause severe degradation when it remains consistent. In contrast, the format-varied stream leads to a large drop for FineTune, whose Last accuracy decreases from $54.1\%$ to $17.3\%$ with \(\Delta_{\mathrm{fmt}}=36.9\). MoELoRA, HiDe-LLaVA, and the semantic-only variant of \ProtoAda\ reduce this gap but still suffer clear format-induced forgetting, with \(\Delta_{\mathrm{fmt}}\) ranging from $3.2$ to $4.7$. Once format-aware prototypes are enabled, the gap is almost eliminated, with \ProtoAda\ reaching $54.7\%$ on the format-varied stream and reducing \(\Delta_{\mathrm{fmt}}\) to only $0.2\). This shows that the observed format mismatch is not only diagnostic but can be effectively mitigated by conditioning task allocation on response protocols.

\begin{figure}[t]
\centering
\includegraphics[width=\linewidth]{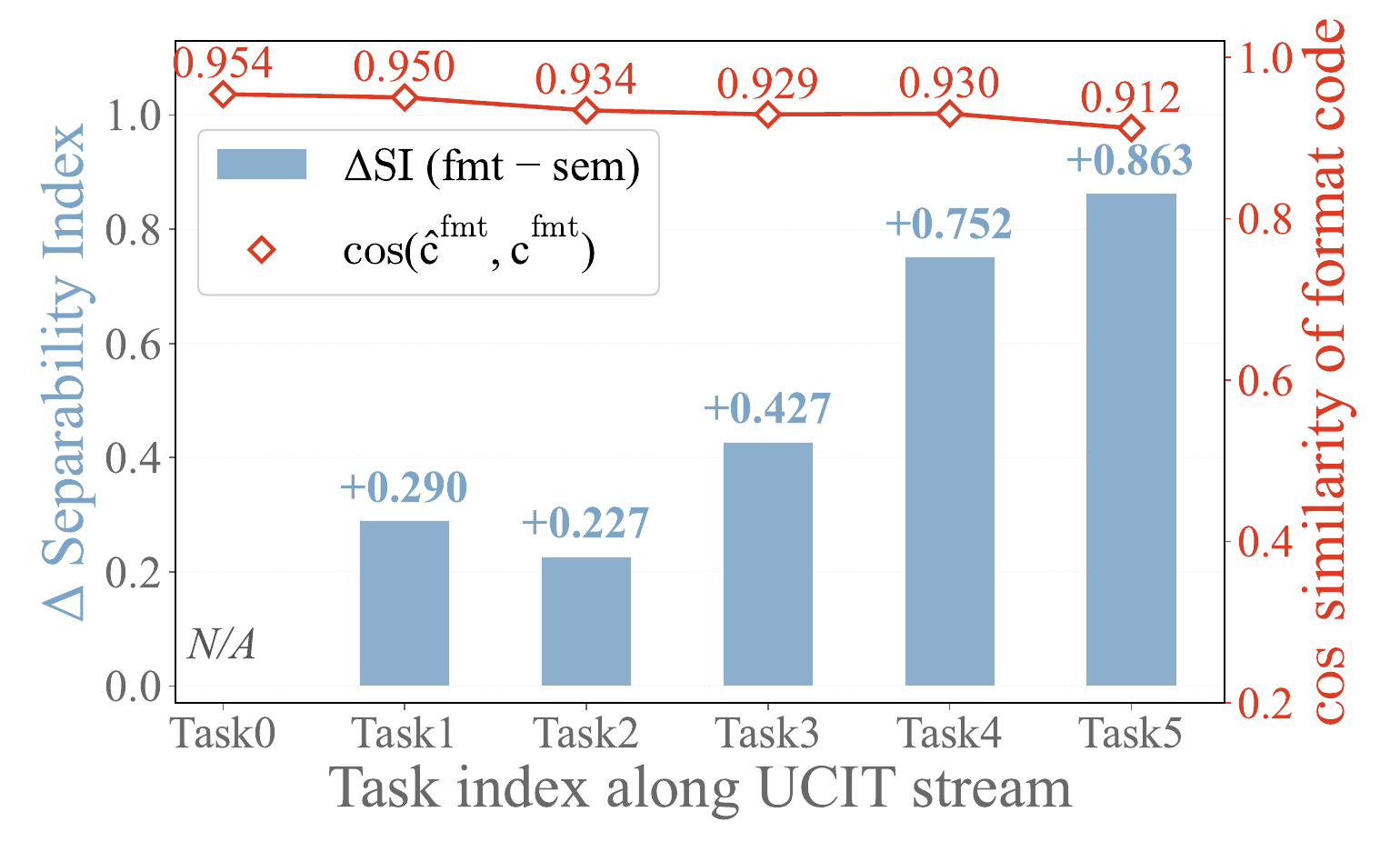}
\caption{Separability index among task prototypes on UCIT as more tasks are learned, with the line tracking similarity between predicted and oracle format codes.}
\label{fig:proto_ev}
    \vspace{-5mm}
\end{figure}

\noindent\paragraph{Temporal Dynamics of Format-Aware Prototypes.}
The visualization in Fig.~\ref{fig:pre_vis_fmt} shows that a lightweight format code separates tasks with different response protocols at a single snapshot, and a natural follow-up question is whether this structure remains stable while new tasks are continuously inserted. Fig.~\ref{fig:proto_ev} tracks this effect along the UCIT stream. Compared with semantic-only prototypes, format-aware prototypes consistently improve the separability index, with \(\Delta \mathrm{SI}\) increasing from $+0.290$ at Task1 to $+0.863$ at Task5. Meanwhile, the predicted format code \(\hat{c}^{\mathrm{fmt}}\) remains highly aligned with the oracle \(c^{\mathrm{fmt}}\), with cosine similarity staying above $0.91$ throughout the stream. These results indicate that the format prototype is not only a useful one-shot descriptor as suggested in Fig.~\ref{fig:pre_vis_fmt}, but also a stable basis for continual task assignment and task-agnostic retrieval.

\begin{figure}[t]
\centering
\includegraphics[width=\linewidth]{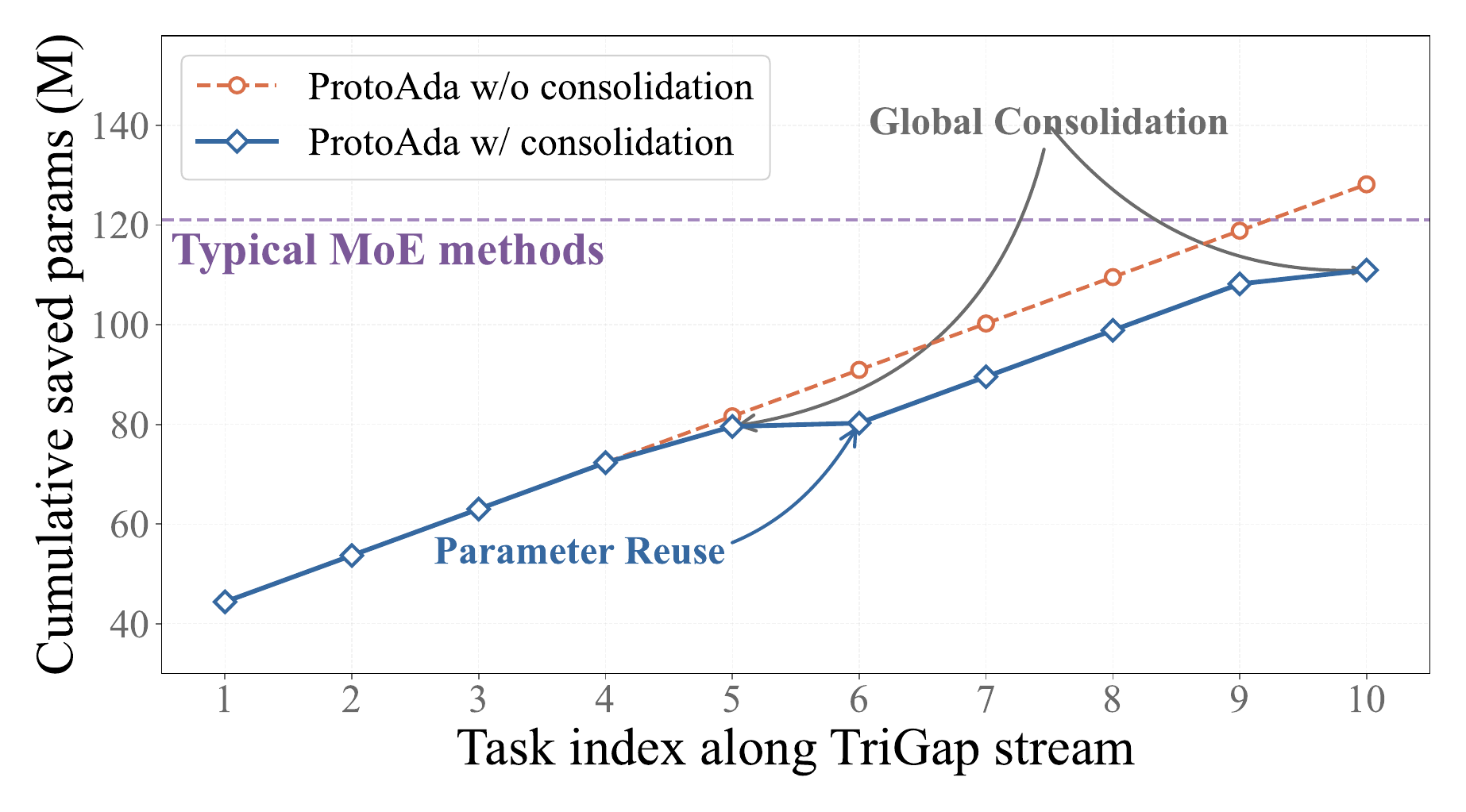}
\caption{Parameter budget of \ProtoAda over the TriGap task stream.}
\label{fig:param_growth}
    \vspace{-4mm}
\end{figure}

\noindent\paragraph{Parameter Growth under Selective Expansion and Consolidation.} 
A core claim of \ProtoAda\ is that selective layer expansion in Eq.~\eqref{eq:layer_selection} and spectral consolidation in Eq.~\eqref{eq:spectral_promotion} keep memory growth controlled rather than letting parameters accumulate linearly. Fig.~\ref{fig:param_growth} records the cumulative saved adapter parameters along the TriGap task stream. Without consolidation, the parameter count grows steadily and reaches about $128$M at Task10, exceeding the typical MoE budget after Task9. In contrast, \ProtoAda\ with consolidation grows more slowly after reuse begins around Task5 and remains below the MoE budget, ending at about $111$M. The widening gap in later tasks shows that global consolidation promotes reusable directions instead of storing every task update independently. Together with the previous analyses, this confirms that \ProtoAda\ improves continual adaptation by reusing compatible directions while preserving task-specific residuals.

\section{Conclusion}
\label{sec:conclusion}
We identify format-blind task assignment as a key source of forgetting in MCIT and propose \ProtoAda, which performs format-aware task grouping, selective LoRA expansion, and geometry-aware update consolidation. By reusing compatible knowledge while preserving task-specific response conventions, \ProtoAda\ achieves stronger continual adaptation across multiple benchmarks.

\section*{Limitations}
Although \ProtoAda{} achieves strong performance on current MCIT benchmarks, its effectiveness has not yet been validated on substantially longer task streams or more diverse response protocols beyond the evaluated settings. Extending \ProtoAda{} to larger-scale continual streams and more open-ended multimodal instruction scenarios remains an important direction for future work.

\bibliography{custom}
\appendix
\clearpage
\section{Theoretical Analysis of Geometry-aware Projection}
\label{app:projection_optimality}

In this section, we provide a theoretical justification for the geometry-aware decomposition in Eq.~\eqref{eq:share_res_decompose}. 
We show that projecting the purified task update onto the group memory subspace extracts the optimal group-compatible component under the Frobenius norm, while the remaining residual captures task-specific directions that are orthogonal to the existing group memory.

\subsection{Preliminaries}

After task $t$ is assigned to group $g_t$, we obtain its purified LoRA update
\begin{equation}
    \Delta \hat{\Theta}_t = \mathbf{M}_t \odot \Delta \Theta_t,
\end{equation}
where $\mathbf{M}_t$ is the activation-aware binary mask. 
Let $\mathcal{U}_{g_t}$ denote the subspace spanned by the current shared memory of group $g_t$. 
Intuitively, $\mathcal{U}_{g_t}$ contains the parameter directions that have been consistently reused by previous tasks in the same group. 
Therefore, an update direction lying in $\mathcal{U}_{g_t}$ can be viewed as group-compatible, while directions outside this subspace are more likely to encode task-specific deviations.

For notational simplicity, we consider the vectorized form of the update:
\begin{equation}
    \mathbf{x}_t = \operatorname{vec}(\Delta \hat{\Theta}_t) \in \mathbb{R}^{d}.
\end{equation}
Let $\mathbf{U}_{g_t} \in \mathbb{R}^{d \times r}$ be an orthonormal basis of $\mathcal{U}_{g_t}$, where $r$ is the dimension of the group subspace and
\begin{equation}
    \mathbf{U}_{g_t}^{\top}\mathbf{U}_{g_t} = \mathbf{I}.
\end{equation}
The orthogonal projection matrix onto $\mathcal{U}_{g_t}$ is then
\begin{equation}
    \mathbf{P}_{g_t} = \mathbf{U}_{g_t}\mathbf{U}_{g_t}^{\top}.
\end{equation}

\subsection{Optimality of the Projected Shared Update}

We first show that the projected component is the closest group-compatible approximation to the purified task update.

\paragraph{Proposition 1.}
Given the purified task update $\Delta \hat{\Theta}_t$ and the group memory subspace $\mathcal{U}_{g_t}$, the projected update
\begin{equation}
    \Delta\Theta_t^{\mathrm{share}}
    =
    \operatorname{proj}_{\mathcal{U}_{g_t}}(\Delta \hat{\Theta}_t)
\end{equation}
is the optimal group-compatible approximation to $\Delta \hat{\Theta}_t$ under the Frobenius norm:
\begin{equation}
    \Delta\Theta_t^{\mathrm{share}}
    =
    \arg\min_{\mathbf{Z}\in \mathcal{U}_{g_t}}
    \left\|
    \Delta \hat{\Theta}_t - \mathbf{Z}
    \right\|_F^2.
    \label{eq:projection_opt_problem}
\end{equation}

\paragraph{Proof.}
Using the vectorized notation, problem~\eqref{eq:projection_opt_problem} is equivalent to
\begin{equation}
    \min_{\mathbf{z}\in \mathcal{U}_{g_t}}
    \left\|
    \mathbf{x}_t - \mathbf{z}
    \right\|_2^2.
\end{equation}
Since $\mathbf{z}\in \mathcal{U}_{g_t}$, there exists a coefficient vector $\mathbf{a}\in \mathbb{R}^{r}$ such that
\begin{equation}
    \mathbf{z} = \mathbf{U}_{g_t}\mathbf{a}.
\end{equation}
Therefore, the optimization becomes
\begin{equation}
    \min_{\mathbf{a}}
    \left\|
    \mathbf{x}_t - \mathbf{U}_{g_t}\mathbf{a}
    \right\|_2^2.
    \label{eq:least_square_projection}
\end{equation}
Taking the derivative of Eq.~\eqref{eq:least_square_projection} with respect to $\mathbf{a}$ and setting it to zero gives
\begin{equation}
    -2\mathbf{U}_{g_t}^{\top}
    \left(
    \mathbf{x}_t - \mathbf{U}_{g_t}\mathbf{a}
    \right)
    = 0.
\end{equation}
Thus,
\begin{equation}
    \mathbf{U}_{g_t}^{\top}\mathbf{x}_t
    =
    \mathbf{U}_{g_t}^{\top}\mathbf{U}_{g_t}\mathbf{a}.
\end{equation}
Because $\mathbf{U}_{g_t}$ is orthonormal, we have
\begin{equation}
    \mathbf{U}_{g_t}^{\top}\mathbf{U}_{g_t} = \mathbf{I}.
\end{equation}
Therefore, the optimal coefficient is
\begin{equation}
    \mathbf{a}^{*}
    =
    \mathbf{U}_{g_t}^{\top}\mathbf{x}_t.
\end{equation}
Substituting $\mathbf{a}^{*}$ back into $\mathbf{z} = \mathbf{U}_{g_t}\mathbf{a}$ yields
\begin{equation}
    \mathbf{z}^{*}
    =
    \mathbf{U}_{g_t}\mathbf{U}_{g_t}^{\top}\mathbf{x}_t
    =
    \mathbf{P}_{g_t}\mathbf{x}_t.
\end{equation}
Returning to the matrix form, this gives
\begin{equation}
    \Delta\Theta_t^{\mathrm{share}}
    =
    \operatorname{proj}_{\mathcal{U}_{g_t}}(\Delta \hat{\Theta}_t).
\end{equation}
Hence, the projection is the minimizer of Eq.~\eqref{eq:projection_opt_problem}.
\hfill $\square$

This result shows that the projection is not an arbitrary decomposition. 
Among all updates that lie in the group memory subspace, it is the one that best reconstructs the current task update.

\subsection{Orthogonality of the Task-specific Residual}

After extracting the group-compatible component, the residual is defined as
\begin{equation}
    \Delta\Theta_t^{\mathrm{res}}
    =
    \Delta \hat{\Theta}_t
    -
    \Delta\Theta_t^{\mathrm{share}}.
\end{equation}
We next show that this residual is orthogonal to all directions in the group subspace.

\paragraph{Proposition 2.}
The residual component satisfies
\begin{equation}
    \left\langle
    \Delta\Theta_t^{\mathrm{res}}, \mathbf{Z}
    \right\rangle_F = 0,
    \qquad
    \forall \mathbf{Z}\in \mathcal{U}_{g_t}.
    \label{eq:residual_orthogonality}
\end{equation}

\paragraph{Proof.}
In vectorized form, the residual is
\begin{equation}
    \mathbf{x}_t^{\mathrm{res}}
    =
    \mathbf{x}_t - \mathbf{P}_{g_t}\mathbf{x}_t
    =
    \mathbf{x}_t - \mathbf{U}_{g_t}\mathbf{U}_{g_t}^{\top}\mathbf{x}_t.
\end{equation}
For any $\mathbf{z}\in \mathcal{U}_{g_t}$, there exists $\mathbf{b}\in \mathbb{R}^{r}$ such that
\begin{equation}
    \mathbf{z} = \mathbf{U}_{g_t}\mathbf{b}.
\end{equation}
Then,
\begin{align}
    \left\langle
    \mathbf{x}_t^{\mathrm{res}}, \mathbf{z}
    \right\rangle
    &=
    \left(
    \mathbf{x}_t - \mathbf{U}_{g_t}\mathbf{U}_{g_t}^{\top}\mathbf{x}_t
    \right)^{\top}
    \mathbf{U}_{g_t}\mathbf{b} \\
    &=
    \mathbf{x}_t^{\top}\mathbf{U}_{g_t}\mathbf{b}
    -
    \mathbf{x}_t^{\top}
    \mathbf{U}_{g_t}
    \mathbf{U}_{g_t}^{\top}
    \mathbf{U}_{g_t}
    \mathbf{b}.
\end{align}
Since $\mathbf{U}_{g_t}^{\top}\mathbf{U}_{g_t}=\mathbf{I}$, we have
\begin{equation}
    \mathbf{U}_{g_t}^{\top}
    \mathbf{U}_{g_t}
    \mathbf{b}
    =
    \mathbf{b}.
\end{equation}
Thus,
\begin{align}
    \left\langle
    \mathbf{x}_t^{\mathrm{res}}, \mathbf{z}
    \right\rangle
    &=
    \mathbf{x}_t^{\top}\mathbf{U}_{g_t}\mathbf{b}
    -
    \mathbf{x}_t^{\top}\mathbf{U}_{g_t}\mathbf{b} \\
    &= 0.
\end{align}
Therefore, the residual is orthogonal to the group memory subspace. 
Returning to the matrix form gives Eq.~\eqref{eq:residual_orthogonality}.
\hfill $\square$

\subsection{Interpretation for Continual Instruction Tuning}

The above results provide a geometric explanation for the proposed consolidation strategy. 
The shared component
\begin{equation}
    \Delta\Theta_t^{\mathrm{share}}
    =
    \operatorname{proj}_{\mathcal{U}_{g_t}}(\Delta \hat{\Theta}_t)
\end{equation}
is the best approximation of the current task update within the group-compatible subspace. 
Therefore, absorbing this component into the group adapter introduces the part of the update that is most consistent with the historical memory of the assigned group.

Meanwhile, the residual
\begin{equation}
    \Delta\Theta_t^{\mathrm{res}}
    =
    \Delta \hat{\Theta}_t
    -
    \Delta\Theta_t^{\mathrm{share}}
\end{equation}
contains directions that cannot be represented by the current group memory. 
Because it is orthogonal to $\mathcal{U}_{g_t}$, directly merging it into the shared group adapter may introduce task-specific deviations into the group-level parameters. 
This is especially harmful in multimodal continual instruction tuning, where different tasks may share similar image-text semantics but require different response protocols.

Therefore, the projection-based decomposition naturally supports the design of \ProtoAda:
\begin{itemize}
    \item the group-aligned component is consolidated into the shared group adapter for parameter reuse;
    \item the orthogonal component is stored separately as a compact residual to preserve task-specific behavior;
    \item the orthogonality between the two parts reduces interference between reusable knowledge and task-specific response conventions.
\end{itemize}

In this sense, the geometry-aware decomposition provides an optimal and interpretable mechanism for extracting group-compatible updates from task-specific LoRA increments.

\section{Algorithmic Description of \ProtoAda}
\label{app:algorithm}
\begin{algorithm}[t]
\small
\caption{ProtoAda Training (per task $t$)}
\label{alg:protoada}
\begin{algorithmic}[1]
\Require Task $\mathcal{T}_t$; frozen backbone $\Theta_0$, encoders $(E_{\mathrm{text}},E_{\mathrm{vis}})$; groups $\{\mathbf{p}_g,\mathbf{A}_g,\mathcal{U}_g\}$; global adapter $\mathbf{A}^{\mathrm{shared}}$; residual bank
\Ensure Updated group adapters, residuals, and format predictor $f_{\mathrm{fmt}}$
\State \textbf{Prototype construction:} Compute $\mathbf{p}_t^{\mathrm{text}}$, $\mathbf{p}_t^{\mathrm{vis}}$, and format code $\mathbf{c}_t^{\mathrm{fmt}}$; fuse into task prototype $\mathbf{q}_t$ (Eqs.~\eqref{eq:text_vis_proto}--\eqref{eq:task_proto}).
\State \textbf{Format-aware assignment:} Compute similarities $s_{t,g}$ (Eq.~\eqref{eq:group_similarity}); assign to $g_t$ or init new group via adaptive threshold (Eq.~\eqref{eq:adaptive_assignment}); update prototype $\mathbf{p}_{g_t}$ (Eq.~\eqref{eq:group_proto_update}).
\State \textbf{Adaptive growth:} Estimate layer sensitivity $\gamma_\ell$ via gradients (Eq.~\eqref{eq:layer_sensitivity}); select top-$M$ layers $\mathcal{S}_t$ using $\gamma_\ell$ and group usage $u_{g_t,\ell}$ (Eq.~\eqref{eq:layer_selection}).
\State \textbf{Task training:} Init LoRA on $\mathcal{S}_t$; optimize with $\mathcal{L}_{\mathrm{ce}}$ to get update $\Delta\Theta_t$; train predictor $\hat{\mathbf{c}}_t^{\mathrm{fmt}} = f_{\mathrm{fmt}} (\mathbf{p}_t^{\mathrm{text}})$ with MSE (Eq.~\eqref{eq:format_predictor}).
\State \textbf{Geometry consolidation:} Purify $\Delta\Theta_t$ with activation norm to get $\Delta \hat{\Theta}_t$ (Eqs.~\eqref{eq:activation_importance}--\eqref{eq:purified_update}); project onto $\mathcal{U}_{g_t}$ to decouple $\Delta\Theta_t^{\mathrm{share}}$ and $\Delta\Theta_t^{\mathrm{res}}$ (Eq.~\eqref{eq:share_res_decompose}).
\State \textbf{Memory update:} Absorb $\Delta\Theta_t^{\mathrm{share}}$ into $\mathbf{A}_{g_t}$ with weight $\lambda_{t,g_t}$ (Eq.~\eqref{eq:group_absorb}); compress $\Delta\Theta_t^{\mathrm{res}}$ via SVD and store as $\mathbf{R}_t$ (Eq.~\eqref{eq:residual_svd}).
\State \textbf{Maintenance (every $K$ tasks):} Promote dominant directions to $\mathbf{A}^{\mathrm{shared}}$ (Eqs.~\eqref{eq:all_groups}--\eqref{eq:spectral_promotion}); prune redundant residuals based on coverage ratio $\rho_t$ (Eq.~\eqref{eq:residual_coverage}).
\end{algorithmic}
\end{algorithm}

\begin{algorithm}[t]
\small
\caption{ProtoAda Inference (Task-agnostic)}
\label{alg:protoada_inference}
\begin{algorithmic}[1]
\Require Input $(v, q)$; frozen backbone $\Theta_0$, encoders $(E_{\mathrm{text}},E_{\mathrm{vis}})$; predictor $f_{\mathrm{fmt}}$; global adapter $\mathbf{A}^{\mathrm{shared}}$; group adapters $\{\mathbf{A}_g\}$; residual bank $\{\mathbf{R}_r\}$
\Ensure Generated response $y$
\State \textbf{Input processing:} Compute descriptors $\mathbf{p}^{\mathrm{text}}$ and $\mathbf{p}^{\mathrm{vis}}$ via frozen encoders; predict format code $\hat{\mathbf{c}}^{\mathrm{fmt}} = f_{\mathrm{fmt}}(\mathbf{p}^{\mathrm{text}})$ (Eq.~\eqref{eq:format_predictor}).
\State \textbf{Query prototype construction:} Fuse visual, linguistic descriptors, and predicted $\hat{\mathbf{c}}^{\mathrm{fmt}}$ to build the query prototype $\mathbf{q}_{\mathrm{query}}$ (Eqs.~\eqref{eq:format_proto}--\eqref{eq:task_proto}).
\State \textbf{Memory retrieval:} Compute similarities $s_g$ between $\mathbf{q}_{\mathrm{query}}$ and group prototypes $\mathbf{p}_g$; retrieve Top-K groups and calculate routing weights $w_g$ (Eq.~\eqref{eq:inference_weight}).
\State \textbf{Residual activation:} Compare $\mathbf{q}_{\mathrm{query}}$ with stored residual prototypes $\mathbf{q}_r$ to retrieve the subset of compatible task-specific residuals $\{\mathbf{R}_r\}$.
\State \textbf{Response generation:} Augment $\Theta_0$ with $\mathbf{A}^{\mathrm{shared}}$, retrieved group adapters (scaled by $w_g$), and selected residuals $\{\mathbf{R}_r\}$; generate response $y$ in a single forward pass.
\end{algorithmic}
\end{algorithm}

Algorithm~\ref{alg:protoada} summarizes the per-task training procedure of \ProtoAda{} in the continual instruction-tuning stream. For each incoming task, \ProtoAda{} first constructs a format-aware prototype from visual semantics, language semantics, and response-format statistics, and uses it to assign the task to a compatible memory group or initialize a new group. After assignment, the method performs prototype-conditioned adaptive growth by combining layer-wise gradient sensitivity with the historical usage pattern of the assigned group, so that LoRA capacity is introduced only at layers that are important for the current task and reusable within the group. The selected LoRA modules are then optimized with the task loss, while the format predictor is trained to infer the response-format code from input-side descriptors for task-agnostic inference. After training, the learned update is purified by activation-aware masking and decomposed according to the group memory geometry: the group-compatible component is absorbed into the group adapter, whereas the remaining task-specific component is compressed and stored as a residual. Periodic maintenance further promotes dominant reusable directions into the global shared adapter and prunes redundant residuals. This procedure enables \ProtoAda{} to reuse compatible knowledge across tasks while preserving task-specific response conventions.

\section{Construction of Format Statistics}
\label{app:fmt}
To incorporate response protocols into task grouping, we derive a task-level format code \(c_t^{\mathrm{fmt}}\) from target responses. 
For each mini-batch, masked label positions are removed and only valid response tokens are used. 
The statistics are organized into four groups: mean of response length, variance of response length, token uncertainty, and template consistency. 
The mean of response length records the average normalized number of target tokens, which distinguishes short-answer, multiple-choice, and long-form generation protocols. 
The variance of response length measures the dispersion of response lengths across samples, reflecting whether the task follows a fixed output template or allows variable-length responses. 
Token uncertainty is measured by the normalized entropy of the empirical token distribution and the unique-token ratio, characterizing how diverse or constrained the target responses are. 
Template consistency is estimated from the repetition ratio and coarse token-id distribution statistics, including the normalized token-id mean and standard deviation as well as low- and middle-band token-id ratios, which serve as tokenizer-independent proxies for option markers, numbers, punctuation, and repeated answer patterns. 
We average these batch-level statistics over the warm-up batches to obtain a fixed-dimensional task-level descriptor $\mathbf{c}_t^{\mathrm{fmt}}$, which is then used for format-aware prototype construction.

\section{Robustness of Task Order}
\begin{table}[t]
\centering
\footnotesize
\caption{Task-order robustness of \textbf{\ProtoAda{}{}} on \textbf{UCIT}. We report average accuracy (\%) under three task sequences.}
\label{tab:task_order_ucit}
\resizebox{\columnwidth}{!}{%
\begin{tabular}{l|ccc}
\hline
\rowcolor{gray!20}
\textbf{\ProtoAda{} (Ours)} & \textbf{Original} & \textbf{Reverse} & \textbf{Random} \\
\hline
Avg. Accuracy (\%) & \textbf{74.66} & \textbf{70.83} & \textbf{71.17} \\
\hline
\end{tabular}%
}
\end{table}

To evaluate whether \ProtoAda{} depends on a favorable task curriculum, we further examine its robustness to task ordering on the UCIT benchmark. 
Specifically, we report the average accuracy under three different task sequences:
\begin{itemize}
    \item \textbf{Original}: the canonical task order defined by the benchmark protocol;
    \item \textbf{Reverse}: the exact reverse of the original sequence, where the last task is presented first;
    \item \textbf{Random}: a randomly shuffled task order. For reproducibility, we fix the random permutation as:
\end{itemize}
\noindent\textbf{UCIT}: Flickr30k $\to$ CLEVR $\to$ ArxivQA $\to$ ImageNet-R $\to$ IconQA $\to$ VizCap.

As shown in Tab.~\ref{tab:task_order_ucit}, \ProtoAda{} maintains stable performance under different task orders, achieving 74.66\%, 70.83\%, and 71.17\% average accuracy under the Original, Reverse, and Random sequences, respectively. 
Although reversing or shuffling the task order introduces a moderate performance drop, \ProtoAda{} still preserves competitive accuracy without relying on the canonical curriculum. 
This robustness mainly benefits from three designs. 
First, \textit{format-aware task grouping} assigns each incoming task according to a prototype that jointly captures visual semantics, language semantics, and response-format statistics, thereby reducing the chance of merging tasks with incompatible instruction protocols. 
Second, \textit{prototype-conditioned adaptive growth} introduces LoRA capacity only in layers that are sensitive to the current task and historically useful for the assigned group, which mitigates unnecessary perturbation to previously learned behaviors. 
Third, \textit{geometry-aware consolidation} separates group-compatible update directions from task-specific residuals, allowing reusable knowledge to be absorbed into group memory while preserving task-specific deviations separately. 
Together, these mechanisms reduce order-dependent interference and make \ProtoAda{} more suitable for continual multimodal instruction tuning under non-stationary task arrivals.

\section{Brief Description of Compared Methods}
\label{sec:compared_methods}

\noindent\textbf{Zero-shot.} 
This baseline directly evaluates the frozen pre-trained LLaVA model on the complete task sequence without updating any parameters. 
It reflects the intrinsic cross-task generalization ability of the original multimodal backbone.

\noindent\textbf{FT-LoRA.} 
FT-LoRA performs continual instruction tuning with standard LoRA adapters. 
For each incoming task, the model is trained sequentially by updating only the LoRA parameters, while the pre-trained backbone remains fixed. 
No explicit mechanism is introduced to mitigate forgetting across tasks.

\noindent\textbf{Replay-LoRA.} 
Replay-LoRA extends sequential LoRA fine-tuning with an episodic memory buffer. 
A small number of samples from previous tasks are stored and mixed with the current task data during training, so that the model can rehearse earlier knowledge while adapting to new instructions.

\noindent\textbf{MoE-LoRA}~\cite{chen2024coin}. 
MoE-LoRA equips the backbone with multiple LoRA experts and learns a routing function to combine their outputs. 
Instead of relying on a single adapter, the model dynamically aggregates expert-specific adaptations, providing additional capacity for heterogeneous tasks in continual tuning.

\noindent\textbf{HiDe-LLaVA}~\cite{guo2025hide}. 
HiDe-LLaVA adopts a task-aware expert decomposition strategy for multimodal continual instruction tuning. 
It maintains task-associated LoRA experts and activates the corresponding expert during training. 
At test time, the task identity is estimated by matching CLIP-based visual and textual anchors, which are then used to select the most relevant expert.

\noindent\textbf{CL-MoE}~\cite{huai2025cl}. 
CL-MoE introduces an input-dependent mixture-of-experts mechanism for continual multimodal instruction tuning. 
It performs fine-grained routing at the layer and token levels, allowing different tokens to access different experts without requiring explicit task identifiers. 
Combined with memory replay, it serves as a strong task-agnostic continual learning baseline.

\noindent\textbf{DISCO}~\cite{guo2025federated}. 
DISCO builds task prototypes from CLIP image and text representations and uses them to guide expert aggregation. 
During inference, the similarity between the current input and stored task prototypes determines diagonal mask weights over LoRA experts, enabling prototype-based routing in the absence of explicit task IDs.

\noindent\textbf{ModalPrompt}~\cite{zeng2025modalprompt}. 
ModalPrompt is a prompt-based continual learning method that assigns learnable soft prompts to different tasks. 
In inference, the model retrieves the most relevant prompts according to dual-modal guidance from image and text features. 
A balancing coefficient controls the relative contribution of visual and textual cues during prompt selection.

\noindent\textbf{SAME}~\cite{xie2026same}. 
SAME exploits spectral information from LoRA parameter updates to identify stable task-related directions. 
It incrementally performs singular value decomposition over recent LoRA statistics and retains dominant spectral anchors for knowledge consolidation. 
A curvature-aware importance measure is further used to regulate parameter updates and reduce interference among tasks.

\end{document}